\documentclass{TEAI}
\usepackage{helvet}
\usepackage[utf8]{inputenc}
\usepackage{array}
\usepackage{float}
\usepackage{url}
\usepackage{wrapfig}
\usepackage{needspace}
\graphicspath{{fig/}}
\newcommand{\roboskill}{\ensuremath{\mathit{RoboSkill}}}
\newcommand{\roboskillminus}{\ensuremath{\mathit{RoboSkill}^{-}}}
\newcommand{\deltacell}[1]{{\scriptsize\color{black!70}#1}}

\title{Explore, Execute, Evolve: A Skill Acquisition and Reuse Loop for Embodied Agents}
\author{
    Sicheng Xie\textsuperscript{1,2,3,*},
    Yitong Chen\textsuperscript{1,2,3,*},
    Haidong Cao\textsuperscript{1},
    Shunlin Lu\textsuperscript{3},\\
    Zuxuan Wu\textsuperscript{1,2,3,$\dagger$},
    Yu-Gang Jiang\textsuperscript{1,$\dagger$}
}

\affiliation{
$^1$\mbox{Institute of Trustworthy Embodied AI, Fudan University}\\
$^2$\mbox{Shanghai Innovation Institute}
\quad
$^3$\mbox{NeoteAI.}
}

\abstract{Vision-language-action and world-action models have demonstrated impressive
capabilities in robotics, yet generalization to unseen tasks remains challenging.
More recently, general-purpose multimodal agents have shown great potential
for zero-shot robotic task solving. However, they often incur high execution costs by reasoning and exploring the physical world from scratch.
To reduce these costs, we introduce
\textbf{RoboSkill}, a framework that connects skill acquisition and reuse
through an \textbf{Explore, Execute, Evolve} loop.
Within this loop, the agent explores to gather task-relevant information,
executes tasks while adapting to feedback, and evolves its skill library
based on execution records. It then reuses these skills to guide exploration
and execution in the next cycle, closing the loop.
To improve loop efficiency, we complement vision with tactile feedback
to reduce uncertainty during physical interaction. We further augment
textual guidance with reusable code to reduce reasoning overhead
during skill reuse.
On LIBERO-10, RoboSkill improves first-episode success rates by
12.5--25.0 percentage points and reduces average runtime by
7.6--72.4\% across four agents. On real robots, it improves success
rates by 8.3 percentage points and reduces average runtime for
successful trials by at least 14.4\%.}

\checkdata[Code]{\url{https://github.com/SII-dannyXSC/RoboSkill}}

\begin{document}
\raggedbottom
\maketitle

\begin{NoHyper}
  \renewcommand{\thefootnote}{}

  \makeatletter
  \renewcommand{\@makefntext}[1]{\noindent#1}
  \makeatother

  \footnotetext{%
    $^{*}$Equal Contribution.\\
    $^{\dagger}$Corresponding authors.%
  }
\end{NoHyper}

\section{Introduction}
\label{sec:introduction}

Embodied AI has advanced rapidly through vision-language-action models~\citep{brohan2023rt,kim2024openvla,pi0,pi0fast,pi05,gr00tn1,wang2026qwen,xvla} and world-action models~\citep{ye2026dreamzero,pai2025mimic,li2026causal,kim2026cosmos,bi2026motus,yuan2026fast}. Although these models support fast task execution, they still face limitations in generalizing to unseen tasks. More recently, general-purpose multimodal
agents~\citep{liang2023code,fu2026cap,zhang2026harness,lu2026aspire}  have begun to exhibit zero-shot task-solving abilities across diverse
robotic tasks. Yet this flexibility remains expensive: each execution largely
starts anew, requiring repeated perception, physical trial and error, and
recovery from failure.

A natural way to avoid starting anew is to reuse skills from prior
executions. Agents in software environments, including web, GUI, and coding
agents, have shown how interaction histories can be distilled into reusable
skills~\citep{shinn2023reflexion,zhao2024expel,wang2023voyager,wang2024agent}. Extending this idea to embodied agents, however, places different
demands on execution. Software interactions typically expose structured states
and relatively reliable action feedback. By contrast, physical interaction
unfolds through partial observations, changing scenes, and embodiment-dependent
constraints. Its outcomes are also uncertain: executing a command does not
guarantee the intended physical effect~\citep{huang2022inner}. Consequently, reusing a skill often
requires more than replaying a fixed plan. The agent must gather missing
information, verify physical outcomes, and recover when execution deviates from
plan.

To connect skill acquisition with subsequent reuse, we introduce \textbf{RoboSkill}, a framework organized around an \textbf{Explore, Execute, Evolve} loop. The agent explores to acquire missing information, executes the task while adapting to feedback, and consolidates execution records into skills. These skills then guide subsequent exploration and execution, whose outcomes support further skill updates. However, applying these skills in a new scene still requires the agent to determine how they apply and translate their guidance into actions. Uncertain observations complicate this adaptation, while textual guidance alone leaves procedural details to be reconstructed through reasoning.

To address these challenges, we emphasize two design choices: tactile feedback during interaction and reusable code within skills. Tactile feedback, broadly including pressure and force/torque signals, complements visual observations with direct evidence of contact and collision~\citep{calandra2018more,zhang2025ta}. For skill reuse, we augment textual guidance with executable code. Text describes strategies and applicability conditions, while code preserves procedures that the agent can adapt to related scenes and tasks~\citep{liang2023code,wang2023voyager,fu2026cap}.

We evaluate RoboSkill with multiple general-purpose agents on LIBERO-10 and real robots. The results demonstrate improvements in task success and execution efficiency, with skills supporting reuse across tasks and agents. Further analyses examine the contributions of tactile feedback and reusable code, as well as the effects of successive skill updates.

\begingroup
\setlength{\intextsep}{4pt}
\setlength{\abovecaptionskip}{5pt}
\begin{figure}[t]
    \centering
    \includegraphics[width=\textwidth]{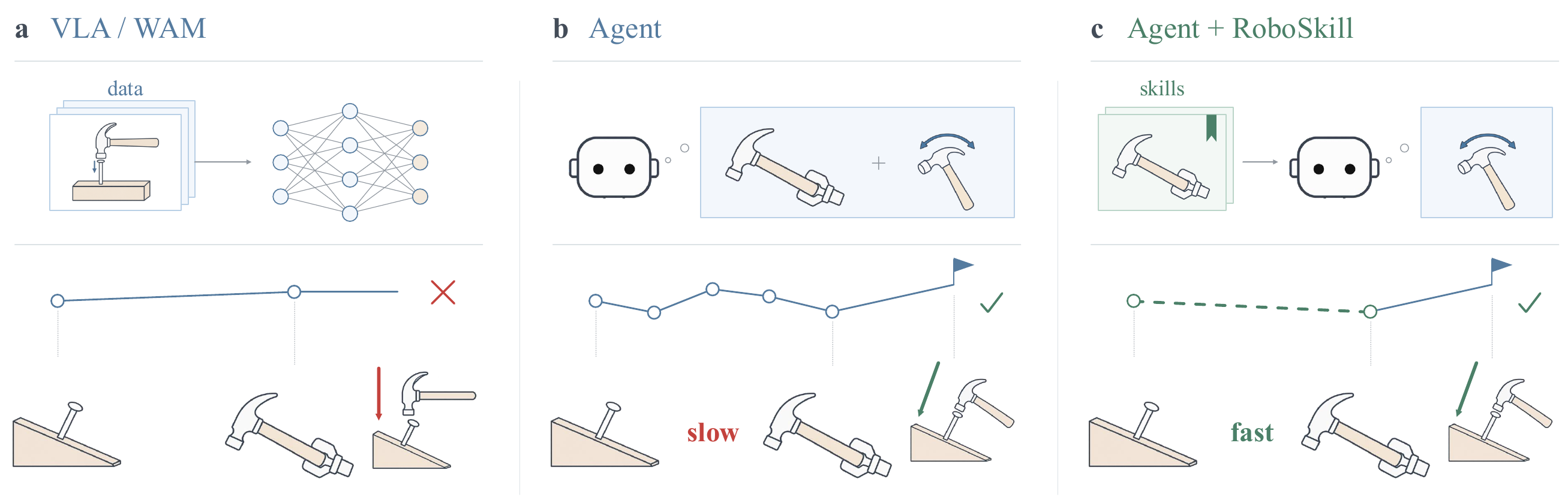}
    \caption{\textbf{RoboSkill accelerates agent execution through skill reuse.} In this tilted-nail example, a VLA/WAM follows the vertical hammering strategy learned from its training data. An embodied agent reasons about both grasping and striking, while RoboSkill accelerates execution by reusing the grasping skill and reasoning only about the strike direction.}
    \label{fig:roboskill-teaser}
\end{figure}
\endgroup

Our contributions are threefold:
\begingroup
\setlength{\leftmargini}{1.2em}
\begin{itemize}
    \item We formulate an Explore, Execute, Evolve loop that connects physical
    interaction with skill construction and reuse. Skills acquired through
    execution guide subsequent exploration and execution, whose outcomes
    inform further updates.

    \item We introduce RoboSkill, a framework that implements this loop through
    the construction, retrieval, adaptation, and update of shared skills.
    Tactile feedback supports physical interaction, while reusable code
    complements textual guidance within skills.

    \item We evaluate RoboSkill in simulation and on real robots, covering
    task performance, cross-task transfer, cross-agent reuse, and successive
    skill updates. Controlled comparisons examine the roles of tactile
    feedback and executable code.
\end{itemize}
\endgroup

\section{Related Work}
\label{sec:related-work}

\paragraph{Vision-Language-Action and World-Action Models}
Vision-language-action (VLA) models have progressed from large-scale multi-task robot policies toward increasingly generalist models trained across diverse tasks, datasets, and embodiments~\citep{brohan2022rt,brohan2023rt,o2024open,team2024octo,kim2024openvla,pi0,pi05}. More recently, world-action models (WAMs) have incorporated predictive world modeling into robot control, using future-state prediction or joint state-action modeling to capture physical dynamics and improve action generation~\citep{pai2025mimic,ye2026dreamzero,li2026causal,bi2026motus,yuan2026fast}. Despite their increasing generality, these models remain largely shaped by the data and tasks encountered during training. RoboSkill explores a complementary direction by leveraging general-purpose agents to reason and adapt during deployment, rather than relying solely on knowledge acquired through policy training.

\paragraph{Language-Model Agents for Robotics.}
Language models have been integrated into robotic systems in several ways. Early approaches use language models to generate executable robot programs or task plans~\citep{liang2023code,singh2022progprompt}, while later methods generate structured spatial representations or constraints that can be converted into robot actions~\citep{huang2023voxposer,huang2024rekep}. More recent work explores several agentic formulations for robotic control. CaP-X~\citep{fu2026cap} uses interactive coding agents that revise robot programs based on execution feedback. Harness VLA~\citep{zhang2026harness} alternates between agent reasoning, analytic primitives, and a frozen VLA for contact-rich execution. RoboClaw~\citep{li2026roboclaw} combines autonomous data collection, policy learning, and long-horizon deployment within an agentic framework. ASPIRE~\citep{lu2026aspire} uses iterative robot exploration to discover, repair, and accumulate executable programs in a reusable skill library. RoboSkill instead emphasizes transferable text-and-code experience that can be retrieved and adapted across tasks and execution agents.

\section{Method}
\label{sec:method}

\begin{figure}[t]
    \centering
    \includegraphics[width=\textwidth]{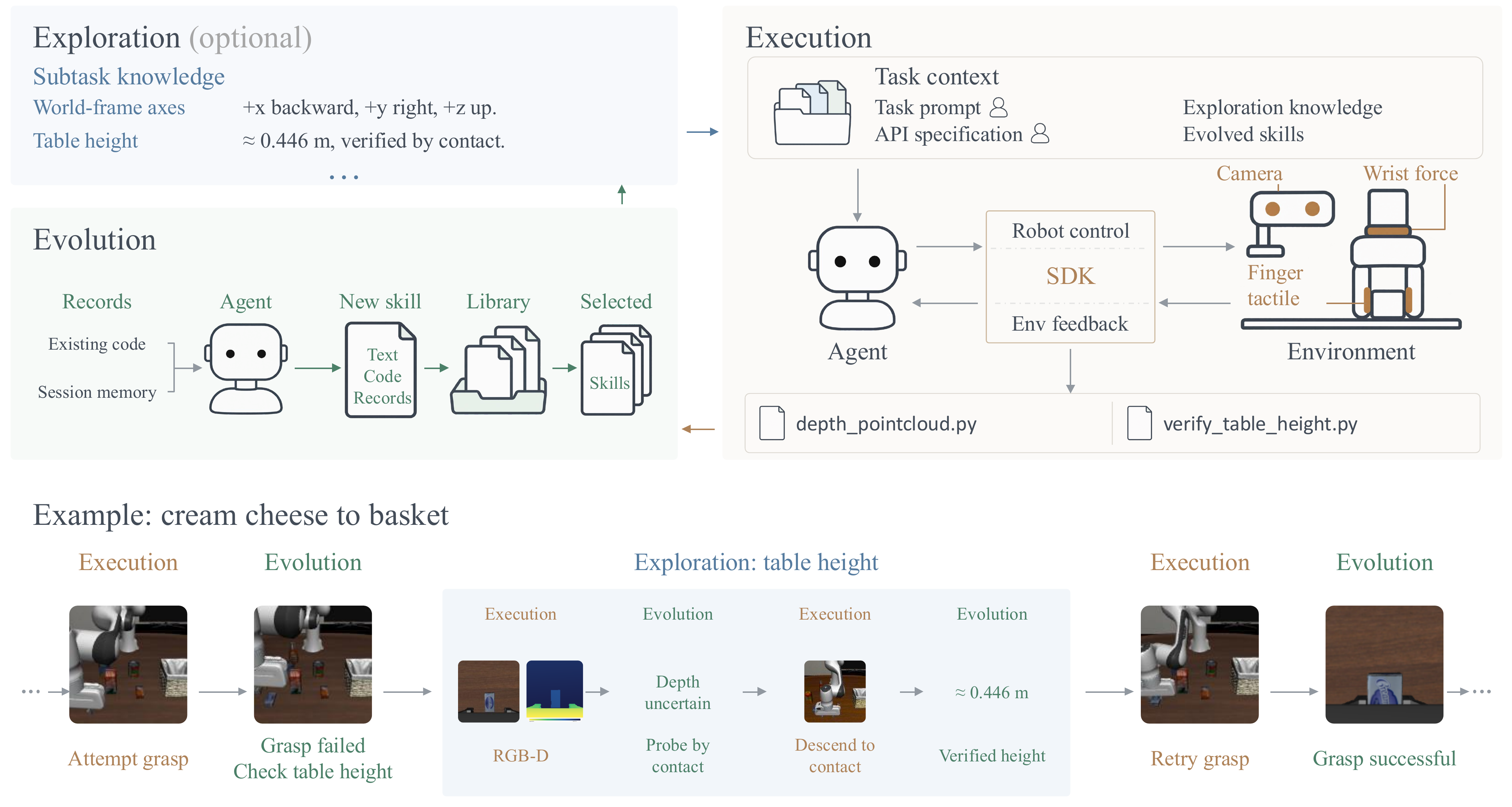}
    \caption{\textbf{The RoboSkill framework.}
The agent explores to acquire missing information, executes with visual and
tactile feedback, and produce skills to guide the next cycle.
The example illustrates a nested exploration loop that verifies table height
through contact before resuming grasping.}
    \label{fig:roboskill-method}
\end{figure}

RoboSkill equips embodied agents with a shared skill library to support an
Explore--Execute--Evolve loop
(\cref{fig:roboskill-method}). 
To describe how this loop supports task completion, we first provide a framework
overview (~\cref{sec:framework-overview}), then introduce exploration
(\cref{sec:method-exploration}) and closed-loop execution
(\cref{sec:method-execution}). Finally, we describe skill evolution (\cref{sec:method-evolution}), which closes the loop by guiding subsequent exploration.

\subsection{Framework Overview}
\label{sec:framework-overview}
To begin the loop, the agent receives a task instruction, selected skills, and access
to the environment through an SDK. When additional information is needed,
the agent solves exploration subtasks to acquire task knowledge. The resulting
knowledge guides the next execution. With this knowledge and the selected
skills, the agent generates and runs code to carry out the task. This code calls
the SDK to perform actions and obtain observations. The agent uses the
resulting feedback to continue or revise execution. When the agent decides
to end execution, it enters the evolution stage and reviews the accumulated
session records and executed code to construct or update skills.
These skills provide guidance for subsequent exploration and execution,
closing the loop from evolution back to exploration.

\subsection{Exploration}
\label{sec:method-exploration}

When task execution requires additional information, we treat acquiring that
information as an exploration subtask. Its immediate goal differs from the
main task, but solving it supports more reliable and efficient completion of
the main task.

\paragraph{Exploration loop.}
We organize exploration subtasks around the same process used to solve the
main task. The agent takes the findings and skills produced by the preceding
evolution as input and uses them to guide further investigation. Solving the
subtask can require multiple rounds of execution and feedback, with each
round informing what to investigate next. The resulting knowledge and
procedures serve as priors for the next execution of the main task.

For example, when moving cream cheese to a basket, a failed grasp can lead
evolution to identify table height as an unresolved issue. The agent then
explores the table height using RGB-D observations. When the depth estimate
remains uncertain, it descends until contact and obtains a height of
approximately $0.446\,\mathrm{m}$. This finding informs the subsequent grasp
attempt. Exploration thus contains its own loop of execution and review
within the main task.

\paragraph{Exploration knowledge.}
We retain exploration findings as current task knowledge in the agent session,
alongside the interaction records and executed code. The agent revises this
knowledge as feedback resolves the exploration subgoal or reveals a need for
further investigation. If the agent exits after completing exploration, the harness
invokes the agent as its own reviewer to consolidate the session records and
executed code into a skill package, following the evolution procedure in
Section~\ref{sec:method-evolution}. If the agent continues in the same session,
the findings remain in session context and directly support the next execution.

\subsection{Closed-loop Execution}
\label{sec:method-execution}

Building on exploration results and available skills, the agent carries out
actions toward the current task and observes their outcomes. Execution may
involve multiple actions and continues until the agent decides to exit.

\paragraph{Context.}
We provide the agent with the task instruction, API specifications for robot
control and sensing, exploration knowledge, and selected
skill packages. This context enables the agent to adapt reusable procedures
to the current scene and compose them into an execution program.

\paragraph{Observations.}
Throughout this paper, \emph{tactile feedback} broadly including pressure and force/torque signals.
We augment visual observations, including RGB and depth,
with tactile feedback to inform the agent about physical interaction.
The agent obtains these observations through API calls during execution.
RGB observations reveal scene state and task progress, depth measurements
support spatial estimation, and tactile feedback provides evidence of contact
that may be difficult to determine from images alone. Together, these
observations support assessment of action outcomes.

\paragraph{SDK-based interaction.}
We provide an SDK instead of tools so that the agent can build reusable code
while interacting with the environment. The agent can compose SDK calls and
data processing into executable procedures. For example, it can create
\texttt{depth\_pointcloud.py} to convert a depth image into a point cloud and
\texttt{verify\_table\_height.py} to verify table height by descending until
contact. These procedures
capture how an operation was performed and can be retained in skill packages
for adaptation and reuse in subsequent tasks.

We record the interactions and executed code throughout execution. When the
agent exits, these records become the input to evolution.

\subsection{Evolution}
\label{sec:method-evolution}

Evolution turns execution records into reusable skills that guide subsequent
exploration and execution. It consolidates the findings and procedures
developed while solving both exploration subtasks and main tasks, closing
the Explore--Execute--Evolve loop.

\paragraph{Evolution timing.}
In principle, each completed subtask should be summarized into a reusable
skill. In practice, invoking a review after every subtask introduces
substantial overhead. We therefore retain findings and execution records in
the session while the task is ongoing, making them available for continued
execution. When the task ends and the agent exits, we consolidate the
accumulated records through a single review.

\paragraph{Skills.}
Each skill contains textual guidance, reusable code, and supporting records.
The textual component summarizes the task
strategy, applicability conditions, and lessons from failed attempts. The
program organizes reusable procedures into a hierarchical structure.
Supporting records include failure cases, full execution trajectories,
and metadata. Failure cases summarize unsuccessful attempts
and corrective strategies, while trajectories preserve the full sequence of
actions and observations. Metadata contains a brief skill description and
visual keyframes to help the agent assess relevance when selecting skills
for a task.

The agent acts as its own reviewer to extract and consolidate these components
from the session records and the code actually executed. If a corresponding
skill already exists, the agent supplements it with the new findings and
procedures; otherwise, it creates a new skill package.

\paragraph{Skill usage.}
To reuse these skills in a new task, the agent inspects their metadata and
selects up to $K$ relevant skill packages. We set $K=3$ in our experiments.
Compared with rule-based selection, agent selection
improves first-attempt success for most evaluated models
(Table~\ref{tab:topk-performance-comparison}). Following selection, we place the selected packages
in the workspace of the agent assigned to the new task, making their guidance
and code available for subsequent exploration and execution.

\section{Experiments}
\label{sec:experiments}

\subsection{Experimental Setup}
\paragraph{Simulation Setup.}
We evaluate on the ten tasks in LIBERO-10~\cite{liu2023libero}.
The main comparison uses four evaluation seeds per task,
excluding the skill-construction seed, for 40 task--seed pairs
(evaluation cells) per condition.
Seed protocols for all experiments are detailed
in~\cref{sec:evaluation-seeds}.
Each cell has a 4 h time budget and may contain multiple episodes:
the initial attempt counts as Episode 1, and each invocation of
\texttt{reset} starts a new episode.
A cell ends when the task succeeds, the agent exits,
or the time budget expires.

\paragraph{Real Robot Setup.}
We conduct experiments across five parallel stations, each equipped
with a Piper robotic arm, tactile sensors on both gripper fingers,
a wrist-mounted camera, and an external overhead camera as shown in~\cref{sec:hardware_setup}. We evaluate on 12 tasks, comprising 6 easy and
6 hard tasks, with 10 trials per task. Task descriptions and
difficulty assignments are provided in~\cref{sec:task_details}.
Each trial consists of a single episode with a 1 h time budget.
A volunteer monitors execution on site and manually determines
task success. Trials end upon success, agent-initiated termination,
or expiration of the time budget.

\paragraph{Agents and Comparison Settings.}
\textbf{Baseline} uses no saved skills.
\textbf{\boldmath\roboskill{}} selects up to three skills from
successful exploration, while \textbf{\boldmath\roboskillminus{}}
uses one same-task skill. Both use textual guidance and executable
code unless otherwise specified. On the real robot, \roboskill{}
uses one same-task skill and is equivalent to \roboskillminus{}.
Simulation skill selections are listed in~\cref{sec:skill_selected}.
We evaluate GPT-6 Astra, GPT-5.6 Sol, Fable 5.1, and Opus 5 in
simulation, and Astra and Sol on the real robot.
All agents use high reasoning or thinking settings and shared
tools through thin CLI adapters. Model parameters, observation
and action interfaces, and time budgets are fixed within each
comparison.

\paragraph{Evaluation Metrics.}
In simulation, we report final success rate (Final SR), success rate
within the first episode (1st SR), and success rate within 30 minutes
(SR@30m), all computed over all evaluation cells. We also report the
mean number of episodes (Avg. Ep.) and mean elapsed time until success
or termination (Avg. Time), both averaged over all cells.
For real-robot experiments, we report Final SR over all trials and
Avg. Time over successful trials only.  Higher success rates and lower
episode counts and times indicate better performance.

\begin{table}[t]
\caption{\textbf{Main results on Libero-10.}}
\label{tab:baseline-roboskill}
\centering
\footnotesize
\setlength{\tabcolsep}{4pt}
\renewcommand{\arraystretch}{1.20}
\begin{tabular*}{\textwidth}{
  @{\extracolsep{\fill}}lccccc@{}
}
\toprule
Model / setting
& \shortstack{Final SR\\$\uparrow$}
& \shortstack{1st SR\\$\uparrow$}
& \shortstack{SR@30m\\$\uparrow$}
& \shortstack{Avg. Ep.\\$\downarrow$}
& \shortstack{Avg. Time\\(min) $\downarrow$} \\
\midrule
\multicolumn{6}{@{}l}{\textbf{GPT-6 Astra}} \\
Baseline
& \textbf{100.0\%} & 72.5\% & 87.5\% & 1.5 & 18.5 \\
\roboskill{}
& \textbf{100.0\%} & \textbf{97.5\%} & \textbf{95.0\%}
& \textbf{1.0} & \textbf{17.1} \\
\midrule
\multicolumn{6}{@{}l}{\textbf{GPT-5.6 Sol}} \\
Baseline
& 82.5\% & 27.5\% & 32.5\% & 9.2 & 94.5 \\
\roboskill{}
& \textbf{97.5\%} & \textbf{52.5\%} & \textbf{72.5\%}
& \textbf{2.9} & \textbf{35.9} \\
\midrule
\multicolumn{6}{@{}l}{\textbf{Fable 5.1}} \\
Baseline
& 97.5\% & 85.0\% & 77.5\% & 2.5 & 34.8 \\
\roboskill{}
& \textbf{100.0\%} & \textbf{97.5\%} & \textbf{95.0\%}
& \textbf{1.1} & \textbf{12.5} \\
\midrule
\multicolumn{6}{@{}l}{\textbf{Opus 5}} \\
Baseline
& \textbf{100.0\%} & 60.0\% & 17.5\% & 1.9 & 80.9 \\
\roboskill{}
& \textbf{100.0\%} & \textbf{80.0\%} & \textbf{82.5\%}
& \textbf{1.4} & \textbf{22.3} \\
\bottomrule
\end{tabular*}
\end{table}

\begin{table}[tbp]
\centering
\begin{minipage}{0.75\textwidth}
\centering
\footnotesize
\caption{\textbf{Main results on real-world using GPT-6 Astra.}}
\label{tab:real-robot-performance}
\setlength{\tabcolsep}{3pt}
\renewcommand{\arraystretch}{1.15}
\begin{tabular*}{\linewidth}{@{\extracolsep{\fill}}lcccc@{}}
\toprule
& \multicolumn{2}{c}{\textbf{Easy Tasks}}
& \multicolumn{2}{c}{\textbf{Hard Tasks}} \\
\cmidrule(lr){2-3}
\cmidrule(l){4-5}
Setting
& \shortstack{1st SR\\$\uparrow$}
& \shortstack{Avg. Time\\(min) $\downarrow$}
& \shortstack{1st SR\\$\uparrow$}
& \shortstack{Avg. Time\\(min) $\downarrow$} \\
\midrule
Baseline
& 91.7\% & 20.3 & 80.0\% & 27.0 \\
\roboskill{}
& \textbf{100.0\%} & \textbf{16.6}
& \textbf{88.3\%} & \textbf{23.1} \\
\bottomrule
\end{tabular*}
\end{minipage}
\end{table}

\subsection{Effectiveness of RoboSkill}
We first evaluate whether reusing skills from successful exploration
improves task completion and execution efficiency.
We compare RoboSkill with Baseline, which requires agents to
solve tasks without saved skills, in both simulation and
real-world experiments.

\roboskill{} improves task success and execution efficiency
in both simulation and real-world experiments.
In simulation, \cref{tab:baseline-roboskill} shows that final
success increases from 82.5\% to 97.5\% for GPT-5.6 Sol
and from 97.5\% to 100.0\% for Fable 5.1, while Astra and
Opus maintain 100.0\% success. All four agents also achieve
higher first-episode success, completing more tasks without resets.

The benefits are especially evident in timely task completion.
For Opus, SR@30m increases from 17.5\% to 82.5\%, and average
execution time decreases from 80.9 to 22.3 minutes.
Sol similarly improves SR@30m from 32.5\% to 72.5\%
and reduces the average episode count from 9.2 to 2.9.
Although Astra shows a smaller reduction in execution time,
its first-episode success rises from 72.5\% to 97.5\%.
These results demonstrate that skill reuse improves efficiency
even when final success is already saturated.

On the real robot, \cref{tab:real-robot-performance} shows
consistent improvements across both difficulty levels.
With Astra, success increases from 91.7\% to 100.0\% on easy
tasks and from 80.0\% to 88.3\% on hard tasks.
Average time over successful trials decreases from 20.3 to
16.6 minutes and from 27.0 to 23.1 minutes, respectively.
Overall, \roboskill{} enables more reliable and efficient
task execution across agents and environments.
\subsection{Cross-Task Skill Transfer}
\paragraph{Transfer with a same-task skill.}
We examine whether supplementing a same-task skill with skills
from other tasks improves execution.
\Cref{tab:same-task-vs-library} compares \roboskillminus{}, which
uses a single same-task skill, with \roboskill{}, which reviews
skill descriptions and selects up to three skills relevant to
the target task. Further details are provided
in~\cref{sec:skill_selected}.

Compared with \roboskillminus{}, \roboskill{} improves
first-episode success for Astra, Fable, and Opus by 5.0,
7.5, and 12.5 percentage points, respectively.
Average execution time decreases for Fable and Opus,
from 13.0 to 12.5 minutes and from 30.4 to 22.3 minutes.
The gains vary across metrics: Astra takes slightly longer
despite higher first-episode success, while Sol reduces
average time from 39.9 to 35.9 minutes but lowers first-episode
success from 67.5\% to 52.5\%.

Overall, \roboskill{} improves first-episode success for three
of the four agents and reduces average execution time for
three of the four agents. These results support the benefit
of allowing agents to select complementary skills beyond
a single same-task skill, while showing that the resulting
trade-off between reliability and speed depends on the agent.
Since skill selection and skill count change together,
the comparison does not isolate the effect of cross-task
content alone.
\begin{table}[t]
\caption{\textbf{Using one skill versus multiple skills on LIBERO-10.}
\roboskillminus{} uses only the skill learned on the target task.
\roboskill{} can additionally select skills learned on other tasks,
using up to three skills in total.}
\label{tab:same-task-vs-library}
\centering
\footnotesize
\setlength{\tabcolsep}{2pt}
\renewcommand{\arraystretch}{1.20}
\begin{tabular*}{\textwidth}{
  @{\extracolsep{\fill}}lcccccccc@{}
}
\toprule
& \multicolumn{2}{c}{\textbf{GPT-6 Astra}}
& \multicolumn{2}{c}{\textbf{GPT-5.6 Sol}}
& \multicolumn{2}{c}{\textbf{Fable 5.1}}
& \multicolumn{2}{c}{\textbf{Opus 5}} \\
\cmidrule(lr){2-3}
\cmidrule(lr){4-5}
\cmidrule(lr){6-7}
\cmidrule(l){8-9}
Setting
& \shortstack{1st SR\\$\uparrow$}
& \shortstack{Avg. Time\\(min) $\downarrow$}
& \shortstack{1st SR\\$\uparrow$}
& \shortstack{Avg. Time\\(min) $\downarrow$}
& \shortstack{1st SR\\$\uparrow$}
& \shortstack{Avg. Time\\(min) $\downarrow$}
& \shortstack{1st SR\\$\uparrow$}
& \shortstack{Avg. Time\\(min) $\downarrow$} \\
\midrule
\roboskillminus{}
& 92.5\% & \textbf{16.5}
& \textbf{67.5\%} & 39.9
& 90.0\% & 13.0
& 67.5\% & 30.4 \\
\roboskill{}
& \textbf{97.5\%} & 17.1
& 52.5\% & \textbf{35.9}
& \textbf{97.5\%} & \textbf{12.5}
& \textbf{80.0\%} & \textbf{22.3} \\
\bottomrule
\end{tabular*}
\end{table}
\begin{table}[tbp]
\centering
\begin{minipage}{0.75\textwidth}
\centering
\footnotesize
\caption{\textbf{Can skills transfer to a different task?}
Other-Task Skill provides one skill learned on a paired source
task, with no skill from the target task.
Baseline uses no skills.}
\label{tab:cross-task-results}
\setlength{\tabcolsep}{3pt}
\renewcommand{\arraystretch}{1.15}
\begin{tabular*}{\linewidth}{@{\extracolsep{\fill}}lcccc@{}}
\toprule
& \multicolumn{2}{c}{\textbf{GPT-5.6 Sol}}
& \multicolumn{2}{c}{\textbf{Opus 5}} \\
\cmidrule(lr){2-3}
\cmidrule(l){4-5}
Setting
& \shortstack{1st SR\\$\uparrow$}
& \shortstack{Avg. Time\\(min) $\downarrow$}
& \shortstack{1st SR\\$\uparrow$}
& \shortstack{Avg. Time\\(min) $\downarrow$} \\
\midrule
Baseline
& 30.0\% & 86.0
& \textbf{60.0\%} & 77.0 \\
Other-Task Skill
& \textbf{44.0\%} & \textbf{66.3}
& 56.0\% & \textbf{43.5} \\
\bottomrule
\end{tabular*}
\end{minipage}
\end{table}

\begin{table}[t]
\caption{\textbf{Cross-agent skill transfer to Kimi K3 on LIBERO-10.}
We evaluate 5 seeds per task, totaling 50 cells per setting.}
\label{tab:transfer-seed5}
\centering
\footnotesize
\setlength{\tabcolsep}{2pt}
\renewcommand{\arraystretch}{1.28}
\begin{tabular}{@{}p{1.33in}*{5}{>{\centering\arraybackslash}p{0.74in}}@{}}
\toprule
Setting & \shortstack{Final SR\\$\uparrow$} & \shortstack{1st SR\\$\uparrow$} & \shortstack{SR@30m\\$\uparrow$} & \shortstack{Avg. Ep.\\$\downarrow$} & \shortstack{Avg. Time\\(min) $\downarrow$} \\
\midrule
Baseline & 48.0\% & 12.0\% & 4.0\% & 3.50 & 180.8 \\
\addlinespace[3pt]
\roboskillminus{} from Astra & 80.0\% & 58.0\% & 20.0\% & \textbf{1.33} & 101.2 \\
\addlinespace[4pt]
\roboskillminus{} from Sol & 84.0\% & 42.0\% & 32.0\% & 3.56 & 86.5 \\
\addlinespace[4pt]
\roboskillminus{} from Fable & 92.0\% & \textbf{66.0\%} & \textbf{60.0\%} & 2.50 & 54.9 \\
\addlinespace[4pt]
\roboskillminus{} from Opus & \textbf{94.0\%} & 58.0\% & 48.0\% & 1.84 & \textbf{52.4} \\
\addlinespace[4pt]
\bottomrule
\end{tabular}
\end{table}

\paragraph{Transfer without a same-task skill.}
\Cref{tab:cross-task-results} evaluates a skill generated on another
task against Baseline. The source--target task pairings
and skill assignment protocol are detailed
in~\cref{sec:cross-task-setup}.
For Sol, first-episode success increases
from 30.0\% to 44.0\%, while average time decreases from 86.0
to 66.3 minutes. For Opus, average time decreases from 77.0
to 43.5 minutes, although first-episode success declines from
60.0\% to 56.0\%. These results show that skills can provide
useful guidance beyond their source tasks, with time savings
for both agents, while improvements in first-episode success
depend on the agent.

\subsection{Cross-Agent Skill Transfer}

\paragraph{Cross-agent transfer in simulator.} \Cref{tab:transfer-seed5} shows that skills from all four source
agents improve Kimi K3 over Baseline under the \roboskillminus{}
setting. First-episode success rises
from 12.0\% to 42.0--66.0\%, while average execution time decreases
from 180.8 minutes to 52.4--101.2 minutes. Skills from Fable yield
the highest first-episode success at 66.0\%, whereas skills from
Opus achieve the shortest average time at 52.4 minutes.
The source agent that provides the highest first-episode success
therefore differs from the one that provides the fastest execution.
These results demonstrate that skills can be reused across agents,
although the magnitude of the benefit depends on the skill source.

\begin{table}[tbp]
\centering
\begin{minipage}{0.75\textwidth}
\centering
\footnotesize
\caption{\textbf{Cross-agent transfer on real-world easy tasks.}
Skills transfer from GPT-6 Astra to GPT-5.6 Sol.}
\label{tab:real-cross-agent}
\setlength{\tabcolsep}{4pt}
\renewcommand{\arraystretch}{1.15}
\begin{tabular*}{\linewidth}{@{\extracolsep{\fill}}lcc@{}}
\toprule
Setting
& \shortstack{1st SR\\$\uparrow$}
& \shortstack{Avg. Time\\(min) $\downarrow$} \\
\midrule
Baseline & 13.3\% & 34.8 \\
Skills from Astra & \textbf{70.0\%} & \textbf{30.5} \\
\bottomrule
\end{tabular*}
\end{minipage}
\end{table}

\paragraph{Cross-agent transfer on the real robot.}
We further evaluate skill transfer from GPT-6 Astra to GPT-5.6 Sol
on six easy real-world tasks, with five trials per task.
Transferred skills increase success from 13.3\% to 70.0\%,
while average time over successful trials decreases from
34.8 to 30.5 minutes. These results extend the evidence for
cross-agent skill reuse to physical execution, with the largest
gain appearing in task success.

\subsection{Skill Evolution}
\begin{table}[!t]
\caption{\textbf{Iterative skill refinement on LIBERO-10.}
Baseline uses no saved skills. \roboskillminus{} uses an initial skill
constructed from successful exploration on the target task.
Each update revises that skill through further successful exploration
by the same model.}
\label{tab:self-evolution}
\centering
\footnotesize
\setlength{\tabcolsep}{3pt}
\renewcommand{\arraystretch}{1.20}
\begin{tabular*}{\textwidth}{
  @{\extracolsep{\fill}}lccccc@{}
}
\toprule
Model / setting
& \shortstack{Final SR\\$\uparrow$}
& \shortstack{1st SR\\$\uparrow$}
& \shortstack{SR@30m\\$\uparrow$}
& \shortstack{Avg. Ep.\\$\downarrow$}
& \shortstack{Avg. Time\\(min) $\downarrow$} \\
\midrule
\multicolumn{6}{@{}l}{\textbf{GPT-5.6 Sol}} \\
Baseline
& 86.0\% & 30.0\% & 38.0\% & 9.0 & 80.3 \\
\roboskillminus{}
& 96.0\% & 58.0\% & 66.0\% & 3.7 & 47.1 \\
\roboskillminus{} (1st update)
& 94.0\% & 58.0\% & 68.0\% & 3.7 & 49.3 \\
\roboskillminus{} (2nd update)
& \textbf{98.0\%} & \textbf{70.0\%} & \textbf{82.0\%}
& \textbf{3.1} & \textbf{30.2} \\
\midrule
\multicolumn{6}{@{}l}{\textbf{Opus 5}} \\
Baseline
& 98.0\% & 64.0\% & 58.0\% & 2.5 & 55.9 \\
\roboskillminus{}
& \textbf{100.0\%} & 78.0\% & 82.0\% & 1.3 & 22.0 \\
\roboskillminus{} (1st update)
& \textbf{100.0\%} & \textbf{82.0\%} & \textbf{86.0\%}
& \textbf{1.3} & \textbf{19.9} \\
\roboskillminus{} (2nd update)
& \textbf{100.0\%} & 80.0\% & 74.0\% & 1.3 & 25.3 \\
\bottomrule
\end{tabular*}
\end{table}
\begin{table}[tbp]
\centering
\begin{minipage}{0.75\textwidth}
\centering
\footnotesize
\caption{\textbf{Skill evolution on real-world hard tasks
using GPT-6 Astra.}}
\label{tab:real-skill-evolution}
\setlength{\tabcolsep}{3pt}
\renewcommand{\arraystretch}{1.15}
\begin{tabular*}{\linewidth}{@{\extracolsep{\fill}}lcc@{}}
\toprule
Setting
& \shortstack{1st SR\\$\uparrow$}
& \shortstack{Avg. Time\\(min) $\downarrow$} \\
\midrule
Baseline & 80.0\% & 27.0 \\
\roboskillminus{} & 88.3\% & \textbf{23.1} \\
\roboskillminus{} (1st update)
& \textbf{93.3\%} & 26.2 \\
\bottomrule
\end{tabular*}
\end{minipage}
\end{table}

\paragraph{Skill evolution in simulation.}
We evaluate successive skill updates using \roboskillminus{}
on seeds 100--104 rather than the
original seeds, with all conditions rerun on the new seed set.
The results therefore differ from those in the main table.
\Cref{tab:self-evolution} shows skill updates can improve
performance beyond the initial skill, but the gains are not
monotonic. For Sol, the first update leaves first-episode success
unchanged and slightly increases average time. The second update
achieves the best results across all metrics, raising
first-episode success from 58.0\% to 70.0\% and reducing average
time from 47.1 to 30.2 minutes relative to the initial skill.

Opus reaches its best performance after the first update, with
first-episode success increasing from 78.0\% to 82.0\% and average
time decreasing from 22.0 to 19.9 minutes. The second update
maintains 100.0\% final success but reduces SR@30m to 74.0\%
and increases average time to 25.3 minutes. Overall, further successful exploration improves skill effectiveness,
although the gains are not monotonic across successive updates.

\paragraph{Skill evolution on the real robot.}
On the six hard real-world tasks as shown in ~\cref{tab:real-skill-evolution}, updating the skill used by
\roboskillminus{} with GPT-6 Astra increases success from
88.3\% to 93.3\%.
However, average time over successful trials rises from 23.1
to 26.2 minutes. The update therefore improves task completion
without improving average successful execution time.
These results demonstrate the value of iterative skill refinement
for improving real-world task completion.

\subsection{Ablation Studies}

\paragraph{Executable Code.}
\Cref{tab:code-ablation} shows that including executable code
generally improves performance over text-only skills.
Code reduces average execution time for all four agents,
while improving or preserving first-episode success for
three of them. Sol benefits the most: first-episode success
increases from 15.0\% to 67.5\%, and average time decreases
from 67.1 to 39.9 minutes. Astra improves both metrics,
while Fable maintains its first-episode success rate
and completes tasks faster. Opus also runs faster, despite
a small decline in first-episode success from 70.0\% to 67.5\%.
Overall, these results favor retaining code alongside textual
guidance, with benefits across most agents.

\begin{table}[!t]
\caption{\textbf{Effect of executable code on LIBERO-10.}
The w/o Code variant removes code from the skill provided to
\roboskillminus{} while retaining its textual guidance.}
\label{tab:code-ablation}
\centering
\footnotesize
\setlength{\tabcolsep}{2pt}
\renewcommand{\arraystretch}{1.20}
\begin{tabular*}{\textwidth}{
  @{\extracolsep{\fill}}lcccccccc@{}
}
\toprule
& \multicolumn{2}{c}{\textbf{GPT-6 Astra}}
& \multicolumn{2}{c}{\textbf{GPT-5.6 Sol}}
& \multicolumn{2}{c}{\textbf{Fable 5.1}}
& \multicolumn{2}{c}{\textbf{Opus 5}} \\
\cmidrule(lr){2-3}
\cmidrule(lr){4-5}
\cmidrule(lr){6-7}
\cmidrule(l){8-9}
Setting
& \shortstack{1st SR\\$\uparrow$}
& \shortstack{Avg. Time\\(min) $\downarrow$}
& \shortstack{1st SR\\$\uparrow$}
& \shortstack{Avg. Time\\(min) $\downarrow$}
& \shortstack{1st SR\\$\uparrow$}
& \shortstack{Avg. Time\\(min) $\downarrow$}
& \shortstack{1st SR\\$\uparrow$}
& \shortstack{Avg. Time\\(min) $\downarrow$} \\
\midrule
\shortstack[l]{\roboskillminus{}\\(w/o Code)}
& 90.0\% & 21.1
& 15.0\% & 67.1
& \textbf{90.0\%} & 16.1
& \textbf{70.0\%} & 32.1 \\
\roboskillminus{}
& \textbf{92.5\%} & \textbf{16.5}
& \textbf{67.5\%} & \textbf{39.9}
& \textbf{90.0\%} & \textbf{13.0}
& 67.5\% & \textbf{30.4} \\
\bottomrule
\end{tabular*}
\end{table}

\paragraph{Tactile Input.}
We evaluate GPT-6 Astra with and without tactile input during
exploration. In simulation, the w/o tactile condition removes
all force-related information and the state of gripper.
On the real robot, it removes tactile-sensor observations.
All other settings remain unchanged.

\Cref{tab:tactile-ablation} shows that tactile input reduces
average execution time from 25.4 to 15.9 minutes, despite
a slight decrease in first-episode success from 82.0\% to 78.0\%.
These results highlight its benefit for execution efficiency.

On the real-robot \textit{Remove Test Tube} task, we conduct
10 trials per condition. Tactile input increases success
from 80.0\% to 90.0\% and reduces average successful execution
time from 33.6 to 20.0 minutes.
These results support the value of tactile input during
exploration for reliable and timely task completion.

\begin{table}[t]
\caption{\textbf{Importance of tactile input during exploration using GPT-6 Astra.} In simulation, we evaluate 5 seeds per task,
totaling 50 cells per setting.}
\label{tab:tactile-ablation}
\centering
\footnotesize
\setlength{\tabcolsep}{3pt}
\renewcommand{\arraystretch}{1.20}
\begin{tabular*}{\textwidth}{
  @{\extracolsep{\fill}}lccccc@{\hspace{12pt}}cc@{}
}
\toprule
& \multicolumn{5}{c}{\textbf{LIBERO-10}}
& \multicolumn{2}{c}{\textbf{Real Robot}} \\
\cmidrule(lr){2-6}
\cmidrule(l){7-8}
Setting
& \shortstack{Final SR\\$\uparrow$}
& \shortstack{1st SR\\$\uparrow$}
& \shortstack{SR@30m\\$\uparrow$}
& \shortstack{Avg. Ep.\\$\downarrow$}
& \shortstack{Avg. Time\\(min) $\downarrow$}
& \shortstack{1st SR\\$\uparrow$}
& \shortstack{Avg. Time\\(min) $\downarrow$} \\
\midrule
With tactile
& \textbf{100\%}
& 78\%
& \textbf{94\%}
& 1.42
& \textbf{15.93}
& \textbf{90\%}
& \textbf{20.00} \\
w.o. tactile
& \textbf{100\%}
& \textbf{82\%}
& 74\%
& \textbf{1.20}
& 25.35
& 80\%
& 33.63 \\
\bottomrule
\end{tabular*}
\end{table}

Together, these results support the Explore--Execute--Evolve
loop of \roboskill{}: physical interaction produces reusable
skills that improve subsequent exploration and execution,
including across tasks and agents. Further executions can
refine skills, while tactile feedback and
code contribute to more effective interaction and reuse.
The benefits observed in simulation and on real robots
demonstrate the value of accumulating and adapting skills
rather than solving each task from scratch.



\setlength{\intextsep}{8pt}
\setlength{\textfloatsep}{12pt plus 2pt minus 2pt}
\setlength{\floatsep}{10pt plus 2pt minus 2pt}
\renewcommand{\topfraction}{0.9}
\renewcommand{\bottomfraction}{0.8}
\renewcommand{\textfraction}{0.08}
\renewcommand{\floatpagefraction}{0.8}
\captionsetup{font=small,skip=6pt}

This supplement describes the implementation of hardware setup (\cref{sec:hardware_setup}) skill reuse
(\cref{sec:appendix-implementation}), experimental protocols
(\cref{sec:appendix-protocols}), detailed results
(\cref{sec:appendix-results}), and skill selection analyses
(\cref{sec:skill_selected}).

\section{Hardware Setup}
\label{sec:hardware_setup}
\begin{figure}[htbp]
    \centering
    \begin{minipage}[c]{0.65\linewidth}
    \includegraphics[width=\linewidth]{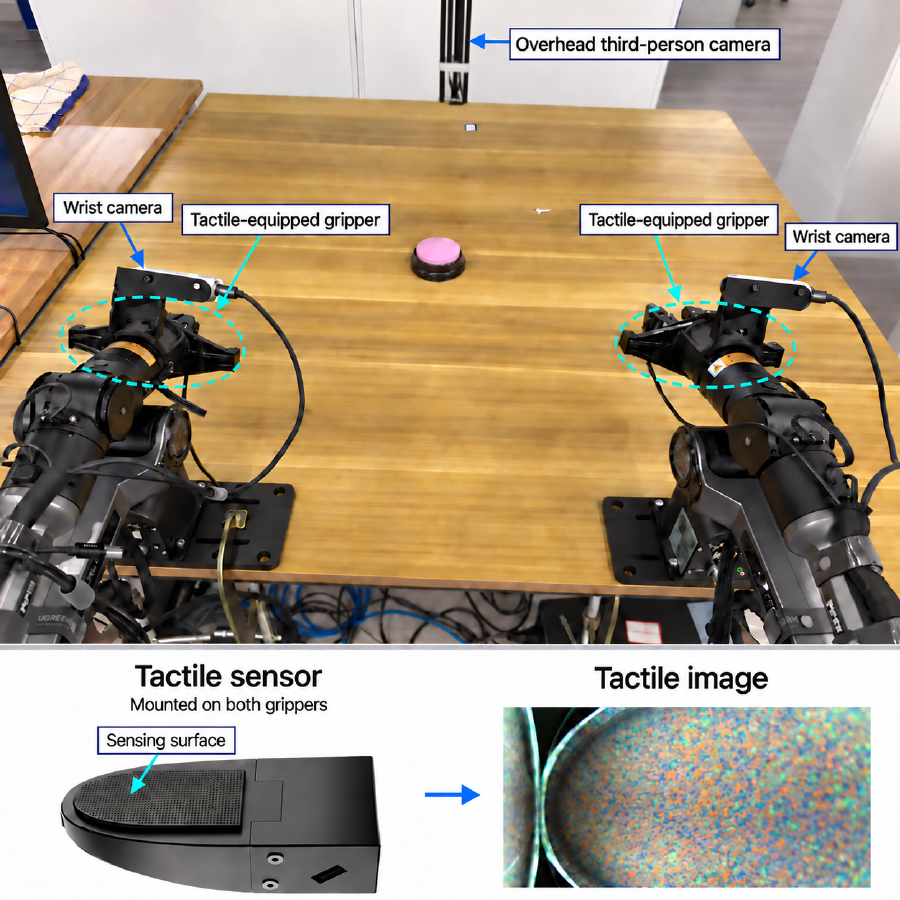}
    \end{minipage}\hfill
    \begin{minipage}[c]{0.32\linewidth}
    \captionsetup{font=small,justification=raggedright,singlelinecheck=false}
    \caption{Real-robot hardware setup.
    Only the right arm is used in the experiments, together with
    its wrist-mounted camera and tactile-equipped gripper.
    The left arm is shown but is not used.
    The upper annotation indicates the support for the overhead
    third-person camera, which is outside the photographed
    field of view. The lower panel illustrates the tactile sensor,
    its sensing surface, and an example tactile image.}
    \label{fig:hardware-setup}
    \end{minipage}
\end{figure}

\Cref{fig:hardware-setup} shows the real-robot hardware setup.
All real-robot experiments use only the right arm, together with
its wrist-mounted camera and tactile-equipped gripper.
The left arm is visible in the photograph but is not used
in the experiments. An external overhead camera provides
a third-person view of the workspace.

\section{Experimental Protocols}
\label{sec:appendix-protocols}

\subsection{Evaluation Seeds}
\label{sec:evaluation-seeds}
We use different seed sets depending on the evaluation objective.
For the main comparison and same-task skill ablations, skills are
constructed using seed 0 and evaluated on seeds 1--4.
We exclude seed 0 from both Baseline and skill-assisted conditions
to measure skill reuse under different initial conditions,
yielding 40 task--seed pairs per setting.

Cross-task and cross-agent transfer experiments use all five
seeds, 0--4, yielding 50 pairs per setting.
Cross-task transfer supplies skills from a different task,
whereas cross-agent transfer supplies skills from a different
agent. The latter evaluates transfer across agents without
excluding the skill-construction seed.
Thus, the 4-seed results measure reuse on held-out initial
conditions, while the 5-seed results cover the full seed set.
\Cref{tab:baseline-seed-protocols} reports Baseline results
under both protocols.

Skill-refinement experiments use a separate evaluation set,
seeds 100--104, with all conditions, including Baseline,
evaluated on these seeds.

\begin{table}[htbp]
\caption{\textbf{Baseline results under the two seed protocols.}
The 4-seed protocol excludes the skill-construction seed and use seed 1-4
for each task; the 5-seed protocol includes all seeds 0--4.}
\label{tab:baseline-seed-protocols}
\centering
\footnotesize
\setlength{\tabcolsep}{4pt}
\renewcommand{\arraystretch}{1.15}
\begin{tabular*}{\textwidth}{@{\extracolsep{\fill}}lccccc@{}}
\toprule
Model
& \shortstack{Final SR\\$\uparrow$}
& \shortstack{1st SR\\$\uparrow$}
& \shortstack{SR@30m\\$\uparrow$}
& \shortstack{Avg. Ep.\\$\downarrow$}
& \shortstack{Avg. Time\\(min) $\downarrow$} \\
\midrule
\multicolumn{6}{@{}l}{\textbf{4 seeds per task: 40 cells}} \\
GPT-6 Astra & 100.0\% & 72.5\% & 87.5\% & 1.50 & 18.48 \\
GPT-5.6 Sol & 82.5\% & 27.5\% & 32.5\% & 9.20 & 94.50 \\
Fable 5.1  & 97.5\% & 85.0\% & 77.5\% & 2.53 & 34.80 \\
Opus 5     & 100.0\% & 60.0\% & 17.5\% & 1.93 & 80.90 \\
\midrule
\multicolumn{6}{@{}l}{\textbf{5 seeds per task: 50 cells}} \\
GPT-6 Astra & 100.0\% & 74.0\% & 90.0\% & 1.46 & 17.82 \\
GPT-5.6 Sol & 86.0\% & 30.0\% & 36.0\% & 7.94 & 86.47 \\
Fable 5.1  & 98.0\% & 84.0\% & 78.0\% & 2.32 & 31.90 \\
Opus 5     & 100.0\% & 60.0\% & 18.0\% & 1.82 & 76.64 \\
\bottomrule
\end{tabular*}
\end{table}

\subsection{Simulation Tasks and Evaluation}
We evaluate on the ten tasks in LIBERO-10. Each evaluation cell is a
task--seed pair with a four-hour budget. The first attempt is Episode 1;
each agent invocation of \texttt{reset} starts a new episode. Evaluation
ends on success, agent-initiated termination, or budget expiration.
Skills are constructed through successful exploration using seeds outside
the evaluation set. The skill evolution experiment uses evaluation seeds
100--104, with all compared conditions rerun on that set.

Final SR, 1st SR, and SR@30m denote success by termination, within the
first episode, and within 30 minutes, respectively. These rates are
computed over evaluation cells. Avg. Ep. and Avg. Time average the number
of episodes and elapsed time until success or termination over all cells,
including failures. Consequently, a shorter Avg. Time alone does not
establish faster successful completion, since an unsuccessful agent can
terminate early.

\begin{figure}[htbp]
    \centering
    \includegraphics[width=0.94\linewidth]{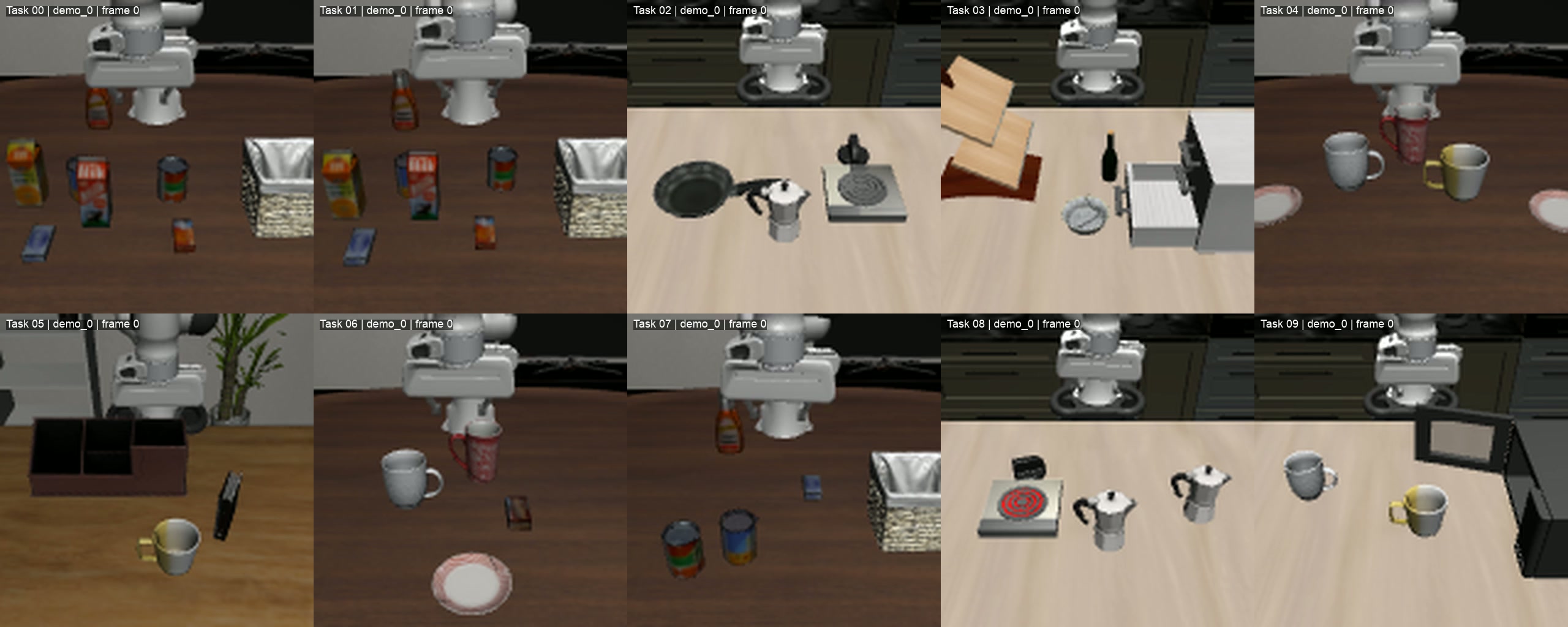}
    \caption{\textbf{Example initial states of the ten LIBERO-10 tasks.}
    Task IDs are used consistently in the cross-task pairings and skill
    selection tables. Each image illustrates one initial state.}
    \label{fig:libero10-tasks}
\end{figure}

\begin{figure}[p]
\centering
\begin{minipage}{\linewidth}
\captionsetup{font=footnotesize,skip=4pt}
\centering
    \includegraphics[width=\textwidth]{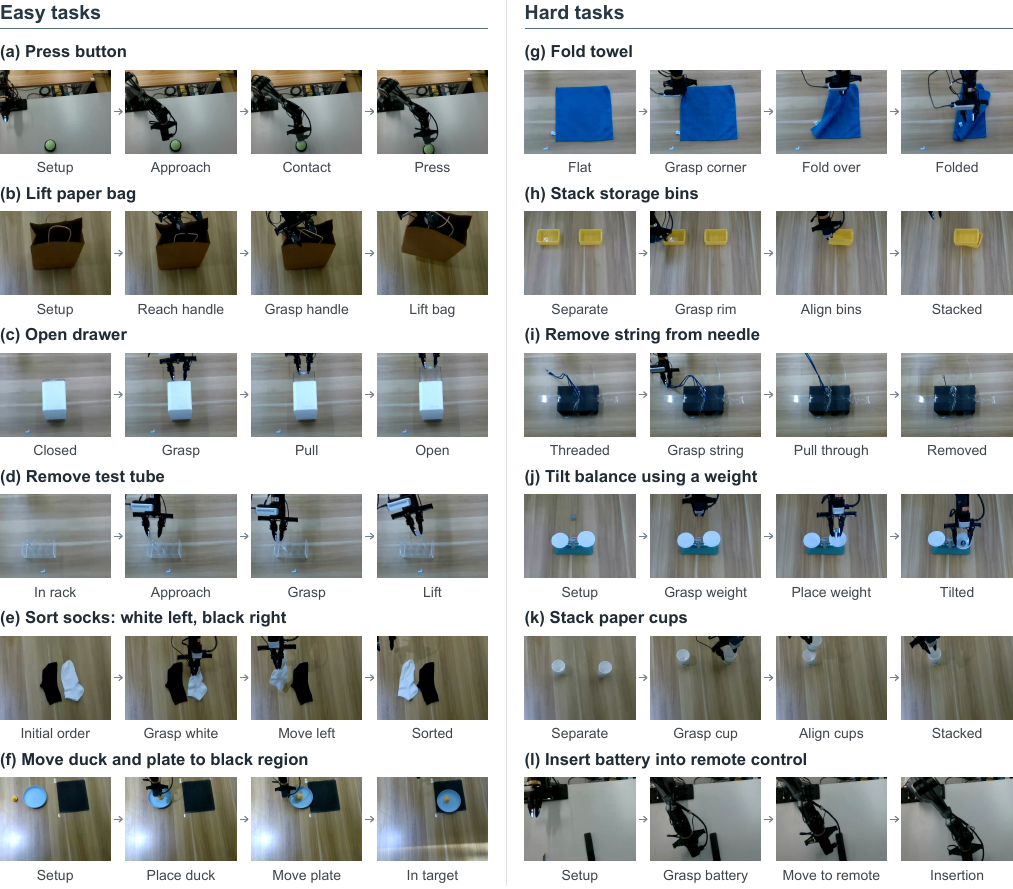}
    \caption{\textbf{Execution stages of the 12 reported real-robot tasks.}
    Panels (a)--(f) show easy tasks and panels (g)--(l) show hard tasks.
    Task objectives are listed in \cref{tab:appendix-task-objectives}.}
    \label{fig:real-robot-task-overview}
\end{minipage}
\par\vspace{10pt}
\begin{minipage}{\linewidth}
\captionsetup{font=footnotesize,skip=4pt}
\centering
\scriptsize
\captionof{table}{\textbf{Real-robot task objectives.}
Letters correspond to the panels in \cref{fig:real-robot-task-overview}.}
\label{tab:appendix-task-objectives}
\setlength{\tabcolsep}{4pt}
\renewcommand{\arraystretch}{1.1}
\begin{tabular}{@{}p{0.05\textwidth}p{0.31\textwidth}p{0.58\textwidth}@{}}
\toprule
ID & Task & Objective \\
\midrule
\multicolumn{3}{@{}l}{\textbf{Easy tasks}} \\
(a) & Press button & Approach and press the button to activate it. \\
(b) & Lift paper bag & Grasp a handle and lift the bag from the table. \\
(c) & Open drawer & Grasp the drawer and pull it outward. \\
(d) & Remove test tube & Grasp a test tube and extract it from the rack. \\
(e) & Sort socks & Arrange white socks on the left and black socks on the right. \\
(f) & Move duck and plate & Relocate both objects into the designated black region. \\
\midrule
\multicolumn{3}{@{}l}{\textbf{Hard tasks}} \\
(g) & Fold towel & Grasp and fold the towel, controlling deformation and alignment. \\
(h) & Stack storage bins & Lift one bin and align it with another to form a stable stack. \\
(i) & Remove string from needle & Grasp the string and guide it out of the needle eye. \\
(j) & Tilt balance using a weight & Place a weight on the balance to change its tilt direction. \\
(k) & Stack paper cups & Grasp one cup and align it with another to nest the cups. \\
(l) & Insert battery into remote control & Grasp the battery, align it with the slot, and insert it. \\
\bottomrule
\end{tabular}
\end{minipage}
\end{figure}

\subsection{Cross-Task and Cross-Agent Transfer}
\label{sec:cross-task-setup}
For cross-task transfer, we use five fixed source--target pairs:
\[
(T_0,T_1),\quad (T_2,T_8),\quad (T_3,T_5),\quad
(T_4,T_9),\quad (T_6,T_7).
\]
Transfer is evaluated in both directions, yielding ten directed
combinations. For example, evaluation on $T_0$ uses a skill constructed
on $T_1$, and evaluation on $T_1$ uses a skill constructed on $T_0$.
The Other-Task Skill condition provides the assigned source skill without
a target-task skill; Baseline provides no saved skills.

For cross-agent transfer in simulation, Kimi K3 receives a single same-task
skill constructed by Astra, Sol, Fable, or Opus. On the real robot,
GPT-5.6 Sol receives skills constructed by GPT-6 Astra and is evaluated
on the six easy tasks, with five trials per task.

\subsection{Real-Robot Tasks and Evaluation}
\label{sec:task_details}
Experiments use five parallel stations, each equipped with a Piper arm,
a wrist-mounted camera, and an external overhead camera. Each trial
contains one episode with a one-hour budget. An on-site volunteer
determines task success. Trials end on success, agent-initiated
termination, or budget expiration. Astra uses ten trials per task;
Sol uses five trials per easy task.

\FloatBarrier
\section{Detailed Experimental Results}
\label{sec:appendix-results}

\subsection{Task-Wise Real-Robot Results}
\label{sec:real-robot-experiments}
\Cref{tab:real-robot-task-performance} gives the per-task Astra results.

\begin{table}[H]
\caption{\textbf{Task-wise Astra performance before and after exploration.}
Baseline timing uses the separate successful-sample records.
\textit{Setting:} 12-task subset (six easy and six hard), ten trials per task
for success rates, 1 h per trial.}
\label{tab:real-robot-task-performance}
\centering
\footnotesize
\setlength{\tabcolsep}{2pt}
\renewcommand{\arraystretch}{1.15}
\begin{tabular*}{\textwidth}{@{\extracolsep{\fill}}p{2.10in}*{4}{c}@{}}
\toprule
Task & \multicolumn{2}{c}{\textbf{Baseline}} & \multicolumn{2}{c}{\textbf{\boldmath\roboskill{}}} \\
\cmidrule(lr){2-3}\cmidrule(l){4-5}
& \shortstack{Final SR\\$\uparrow$} & \shortstack{Avg. Time\\(min) $\downarrow$} & \shortstack{Final SR\\$\uparrow$} & \shortstack{Avg. Time\\(min) $\downarrow$} \\
\midrule
\multicolumn{5}{@{}l}{\textbf{Easy tasks}} \\
Press button & \textbf{10/10} & 17.68 & \textbf{10/10} & \textbf{9.10} \\
Lift paper bag & \textbf{10/10} & 19.58 & \textbf{10/10} & \textbf{15.70} \\
Open drawer & 8/10 & 22.53 & \textbf{10/10} & \textbf{15.40} \\
Remove test tube & 9/10 & 20.00 & \textbf{10/10} & \textbf{16.00} \\
Sort white socks left and black socks right & \textbf{10/10} & \textbf{20.92} & \textbf{10/10} & 21.40 \\
Move duck and plate to black region & 8/10 & 21.93 & \textbf{10/10} & \textbf{21.90} \\
\addlinespace[2pt]
\midrule
\multicolumn{5}{@{}l}{\textbf{Hard tasks}} \\
Fold towel & \textbf{10/10} & 24.02 & \textbf{10/10} & \textbf{23.50} \\
Stack storage bins & 9/10 & \textbf{18.88} & \textbf{10/10} & 19.60 \\
Remove string from needle & \textbf{7/10} & 37.73 & 5/10 & \textbf{37.40} \\
Tilt balance using a weight & 8/10 & 23.15 & \textbf{10/10} & \textbf{16.50} \\
Stack paper cups & 7/10 & 32.63 & \textbf{9/10} & \textbf{24.00} \\
Insert battery into remote control & 7/10 & 29.83 & \textbf{9/10} & \textbf{25.22} \\
\midrule
Overall & 85.8\% & 23.29 & \textbf{94.2\%} & \textbf{19.65} \\
\bottomrule
\end{tabular*}
\end{table}

\Cref{tab:real-robot-gpt56-task-performance} reports the six easy-task
results for transfer from Astra to Sol. Transferred skills enable
successes on all six tasks, including three tasks with no recorded
Baseline successes. 
\begin{table}[H]
\caption{\textbf{Task-wise GPT-5.6 Sol performance before and after receiving
skills constructed by Astra.} Avg. Time uses successful trials only.
\textit{Setting:} six easy tasks, five trials per task, tactile input, and a
1 h limit per trial.}
\label{tab:real-robot-gpt56-task-performance}
\centering
\footnotesize
\setlength{\tabcolsep}{2pt}
\renewcommand{\arraystretch}{1.15}
\begin{tabular*}{\textwidth}{@{\extracolsep{\fill}}p{2.10in}*{4}{c}@{}}
\toprule
Task & \multicolumn{2}{c}{\textbf{Baseline}} &
\multicolumn{2}{c}{\textbf{\boldmath\roboskill{} from Astra}} \\
\cmidrule(lr){2-3}\cmidrule(l){4-5}
& \shortstack{Final SR\\$\uparrow$} & \shortstack{Avg. Time\\(min) $\downarrow$}
& \shortstack{Final SR\\$\uparrow$} & \shortstack{Avg. Time\\(min) $\downarrow$} \\
\midrule
Press button & 0/5 & -- & \textbf{5/5} & \textbf{25.40} \\
Lift paper bag & 2/5 & 27.00 & \textbf{5/5} & \textbf{19.60} \\
Open drawer & 1/5 & \textbf{27.00} & \textbf{3/5} & 34.33 \\
Remove test tube & 1/5 & 58.00 & \textbf{3/5} & \textbf{35.67} \\
Sort white socks left and black socks right & 0/5 & -- & \textbf{2/5} & \textbf{29.00} \\
Move duck and plate to black region & 0/5 & -- & \textbf{3/5} & \textbf{49.00} \\
\midrule
Overall & 13.3\% & 34.75 & \textbf{70.0\%} & \textbf{30.48} \\
\bottomrule
\end{tabular*}
\end{table}


\FloatBarrier
\section{Skill Selection Analysis}
\label{sec:skill_selected}

\subsection{Selected Skills}
\Cref{tab:topk-selection-comparison} compares a fixed task mapping with
agent-selected skill sets. Each entry lists source-task IDs in selection
order. The agent may select up to three skills. All four models retain
the same-task skill as the first entry for every target task; the observed
selection behavior therefore augments a same-task skill with skills from
other tasks.

\begin{table}[H]
\caption{\textbf{Agent-selected versus rule-based Top-K skill sets.}
Each cell lists the source-task skills selected for the target task.
Bold agent-selected cells differ from the fixed rule in membership or order.
}
\label{tab:topk-selection-comparison}
\centering
\scriptsize
\setlength{\tabcolsep}{2pt}
\renewcommand{\arraystretch}{1.15}
\begin{tabular*}{\textwidth}{@{\extracolsep{\fill}}p{1.30in}*{5}{p{0.74in}}@{}}
\toprule
Target task & Rule & GPT-5.6 Sol & Opus 5 & Fable 5.1 & GPT-6 Astra \\
\midrule
T0: Two cans to basket
  & \texttt{[T0,T7,T1]} & \texttt{[T0,T7,T1]} & \texttt{[T0,T7,T1]}
  & \texttt{[T0,T7,T1]} & \textbf{\texttt{[T0,T7]}} \\
T1: Cheese and butter to basket
  & \texttt{[T1,T7,T0]} & \textbf{\texttt{[T1,T7]}} & \texttt{[T1,T7,T0]}
  & \texttt{[T1,T7,T0]} & \textbf{\texttt{[T1,T7]}} \\
T2: Turn on stove; place moka pot
  & \texttt{[T2,T8]} & \texttt{[T2,T8]} & \texttt{[T2,T8]}
  & \texttt{[T2,T8]} & \texttt{[T2,T8]} \\
T3: Black bowl to drawer; close
  & \texttt{[T3,T9]} & \textbf{\texttt{[T3,T9,T5]}} & \texttt{[T3,T9]}
  & \texttt{[T3,T9]} & \textbf{\texttt{[T3]}} \\
T4: Two cups to left/right plates
  & \texttt{[T4,T6,T9]} & \textbf{\texttt{[T4,T6]}} & \textbf{\texttt{[T4,T6]}}
  & \textbf{\texttt{[T4,T6]}} & \textbf{\texttt{[T4,T6]}} \\
T5: Book to rear caddy slot
  & \texttt{[T5]} & \textbf{\texttt{[T5,T3]}} & \textbf{\texttt{[T5,T7]}}
  & \textbf{\texttt{[T5,T3]}} & \texttt{[T5]} \\
T6: White cup to plate; pudding right
  & \texttt{[T6,T4,T0]} & \textbf{\texttt{[T6,T4]}} & \textbf{\texttt{[T6,T4]}}
  & \textbf{\texttt{[T6,T4]}} & \textbf{\texttt{[T6,T4]}} \\
T7: Soup can and cheese to basket
  & \texttt{[T7,T1,T0]} & \textbf{\texttt{[T7,T0,T1]}} & \texttt{[T7,T1,T0]}
  & \textbf{\texttt{[T7,T0,T1]}} & \texttt{[T7,T1,T0]} \\
T8: Two moka pots to stove
  & \texttt{[T8,T2,T0]} & \textbf{\texttt{[T8,T2]}} & \textbf{\texttt{[T8,T2]}}
  & \textbf{\texttt{[T8,T2]}} & \textbf{\texttt{[T8,T2]}} \\
T9: Cup to microwave; close
  & \texttt{[T9,T3,T4]} & \textbf{\texttt{[T9,T4,T3]}} & \textbf{\texttt{[T9,T3]}}
  & \texttt{[T9,T3,T4]} & \textbf{\texttt{[T9]}} \\
\bottomrule
\end{tabular*}
\end{table}

All four models omit the third rule-selected skill for T4, T6, and T8.
The mean number of selected skills decreases from 2.6 under the fixed
rule to 2.4, 2.3, 2.4, and 1.8 for Sol, Opus, Fable, and Astra,
respectively. Astra selects only the same-task skill for T3, T5, and T9.
Selection also adds skills: for T5, Sol and Fable include T3, while Opus
includes T7, although the fixed rule includes only T5.

\subsection{Selection Performance and Interpretation}
\Cref{tab:topk-performance-comparison} compares the resulting performance.
Agent selection improves first-episode success for Astra, Fable, and
Opus, but reduces it for Sol. Average time decreases for Astra and Opus
and increases for Sol and Fable. Agent selection therefore does not
dominate the fixed rule across all models and metrics.

\begin{table}[H]
\caption{\textbf{Rule-based versus agent-selected Top-K performance.}
$\Delta$ is the change from rule-based to agent-selected Top-K (percentage
points for rates and relative percent otherwise). Bold marks the better
non-tied value within each model and metric. }
\label{tab:topk-performance-comparison}
\centering
\scriptsize
\setlength{\tabcolsep}{2pt}
\renewcommand{\arraystretch}{1.0}
\begin{tabular}{@{}p{1.33in}*{5}{>{\centering\arraybackslash}p{0.74in}}@{}}
\toprule
Model / setting & \shortstack{Final SR\\$\uparrow$} &
\shortstack{1st SR\\$\uparrow$} & \shortstack{SR@30m\\$\uparrow$} &
\shortstack{Avg. Ep.\\$\downarrow$} & \shortstack{Avg. Time\\(min) $\downarrow$} \\
\midrule
\multicolumn{6}{@{}l}{\textbf{GPT-6 Astra}} \\
Rule-based Top-K & 100\% & 87.5\% & 85\% & 1.175 & 19.2 \\
Agent-selected Top-K & 100\% & \textbf{97.5\%} & \textbf{95\%} & \textbf{1.025} & \textbf{17.1} \\[-6pt]
\deltacell{$\Delta$ agent vs. rule} & \deltacell{+0 pp} & \deltacell{\textbf{+10 pp}} & \deltacell{\textbf{+10 pp}} & \deltacell{\textbf{-12.8\%}} & \deltacell{\textbf{-11.2\%}} \\
\addlinespace[2pt]
\midrule
\multicolumn{6}{@{}l}{\textbf{GPT-5.6 Sol}} \\
Rule-based Top-K & 97.5\% & \textbf{57.5\%} & 70\% & 3.425 & \textbf{35.4} \\
Agent-selected Top-K & 97.5\% & 52.5\% & \textbf{72.5\%} & \textbf{2.925} & 35.9 \\[-6pt]
\deltacell{$\Delta$ agent vs. rule} & \deltacell{+0 pp} & \deltacell{-5 pp} & \deltacell{\textbf{+2.5 pp}} & \deltacell{\textbf{-14.6\%}} & \deltacell{+1.2\%} \\
\addlinespace[2pt]
\midrule
\multicolumn{6}{@{}l}{\textbf{Fable 5.1}} \\
Rule-based Top-K & 100\% & 95\% & \textbf{97.5\%} & \textbf{1.050} & \textbf{11.2} \\
Agent-selected Top-K & 100\% & \textbf{97.5\%} & 95\% & 1.075 & 12.5 \\[-6pt]
\deltacell{$\Delta$ agent vs. rule} & \deltacell{+0 pp} & \deltacell{\textbf{+2.5 pp}} & \deltacell{-2.5 pp} & \deltacell{+2.4\%} & \deltacell{+12.3\%} \\
\addlinespace[2pt]
\midrule
\multicolumn{6}{@{}l}{\textbf{Opus 5}} \\
Rule-based Top-K & 100\% & 62.5\% & 77.5\% & 1.825 & 27.1 \\
Agent-selected Top-K & 100\% & \textbf{80\%} & \textbf{82.5\%} & \textbf{1.375} & \textbf{22.3} \\[-6pt]
\deltacell{$\Delta$ agent vs. rule} & \deltacell{+0 pp} & \deltacell{\textbf{+17.5 pp}} & \deltacell{\textbf{+5 pp}} & \deltacell{\textbf{-24.7\%}} & \deltacell{\textbf{-17.9\%}} \\
\bottomrule
\end{tabular}
\end{table}

The selection lists document which packages are made available to the
agent. They do not establish which code or textual components are
subsequently used, or the contribution of each selected skill. Since
skill count and membership can change together, this comparison also
does not isolate the effect of skill count from that of skill relevance.

\FloatBarrier

\section{Conclusion}
We presented \textbf{RoboSkill}, a framework that connects skill acquisition
and reuse through an Explore, Execute, Evolve loop. The agent reuses skills
to guide subsequent exploration and execution, then updates them from new
outcomes. Tactile feedback reduces interaction uncertainty, while reusable
code reduces reasoning overhead during skill reuse. Experiments on LIBERO-10
and real robots demonstrate improved success rates and execution efficiency,
with skills supporting reuse across tasks and agents.

\bibliographystyle{plainnat}
\bibliography{references}

\end{document}